%% file: main.tex
\documentclass[10pt,twocolumn,letterpaper]{article}

\usepackage[pagenumbers]{cvpr} 
\usepackage{placeins}
\usepackage{threeparttable}


\definecolor{cvprblue}{rgb}{0.21,0.49,0.74}
\usepackage[pagebackref,breaklinks,colorlinks,allcolors=cvprblue]{hyperref}

\def\paperID{*****}
\def\confName{CVPR}
\def\confYear{2027}

\title{TaskGuard: Task-Conditioned Restoration Utility for Risk-Aware Object Detection}

\author{
Van Vung Pham\\
Computer Science Department, Sam Houston State University, Huntsville, Texas, USA\\
{\tt\small vung.pham@shsu.edu}
\and
}

\begin{document}
\maketitle

\begin{abstract}
Image restoration is commonly applied before object detection under adverse
conditions, yet a visually improved image need not improve the downstream
task.
We study this mismatch as \emph{restoration utility prediction}: given a
degraded image and its candidate restoration, should the restoration be used
or should the original observation be preserved?
We introduce \textbf{TaskGuard}, a post-hoc controller for frozen restoration
and detection pipelines.
TaskGuard characterizes the realized restoration residual through its
interaction with detector sensitivity and predicts whether the intervention is
task-beneficial.
Exact regional counterfactuals reveal substantial within-image utility
heterogeneity, while a deployable pseudo-gradient preserves statistically
reliable directional information.
Feature-group ablation further shows that task-conditioned evidence contributes
information beyond detector-response and residual statistics.
The TaskGuard utility predictor is trained only on Gaussian degradation and
frozen before final evaluation, then transferred to unseen motion blur, rain,
and defocus.
Across these unseen families, TaskGuard reduces loss-negative interventions by
\textbf{54.2\%} (family macro) and practical per-image detection deteriorations
by \textbf{37.0\%} (pooled), while preserving \textbf{98.8\%} of the
Always-Restore COCO AP.
On natural-rain DAWN, it reduces loss-negative interventions by
\textbf{97.9\%} while retaining \textbf{77.8\%} of the AP improvement obtained
by deraining.
These results support restoration utility as a task-conditioned property of
the specific intervention rather than image appearance alone.
\end{abstract}

\section{Introduction}
\label{sec:introduction}

Image degradation caused by noise, blur, rain, and other adverse conditions can
substantially reduce object-detection accuracy, motivating restoration before
recognition
~\cite{liu2022iayolo,kalwar2023gdip,li2023detectionfriendly}.
Modern restoration methods can recover considerable aggregate detection
performance, but their effects are not uniformly beneficial.
A perceptually cleaner image may remove, alter, or introduce evidence important
to the downstream model
~\cite{kim2024sr4ir,chen2025unirestore,li2025machinepreference}.
Consequently, unconditional restoration may improve average accuracy while
making individual observations worse~\cite{fang2026lqtaranker}.

\begin{figure*}[t]
    \centering
    \includegraphics[width=\textwidth]
    {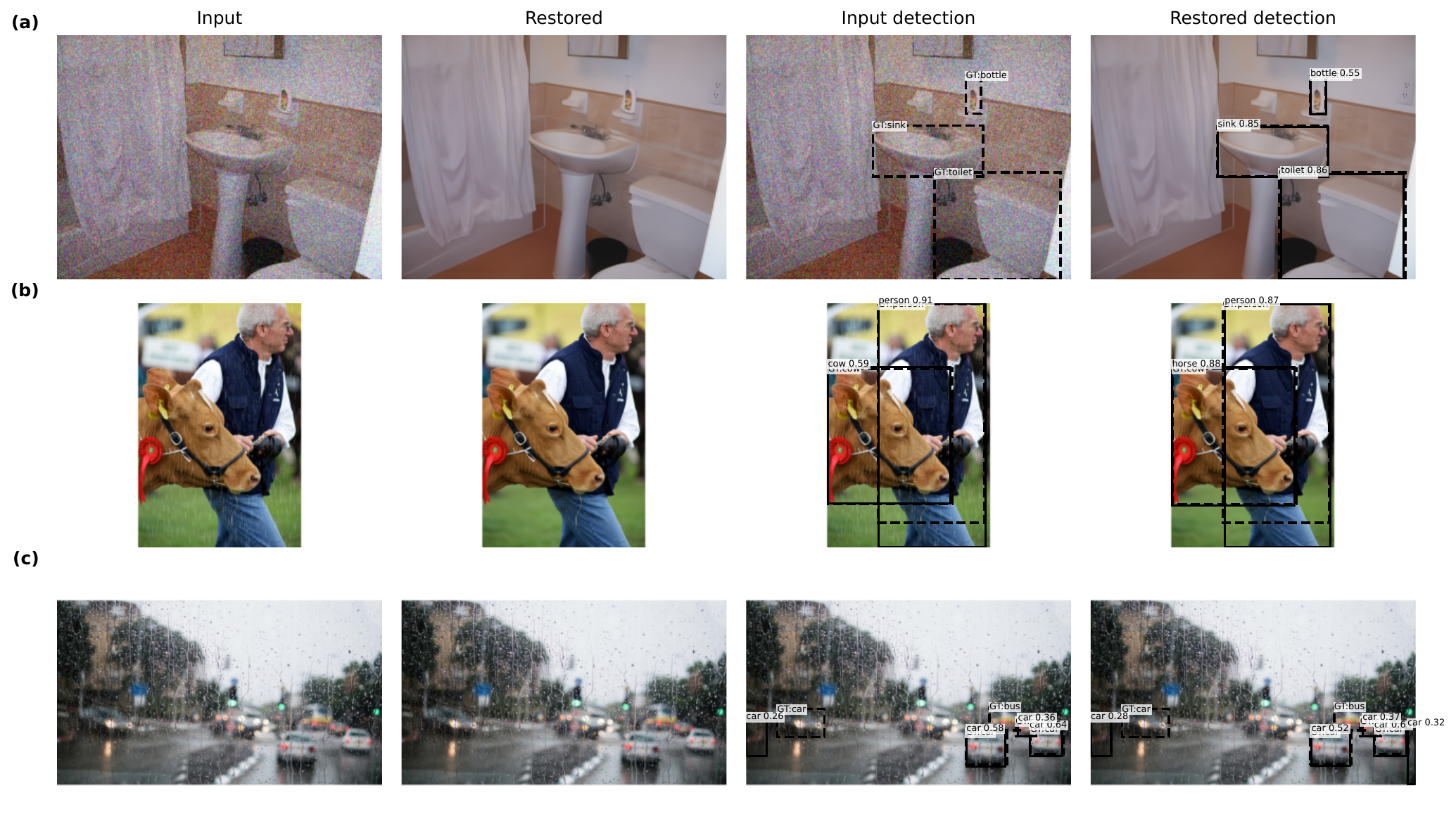}
    \caption{
    Examples of restoration utility.
    \textbf{(a)} Gaussian restoration recovers three missed objects.
    \textbf{(b)} Rain restoration changes a correct \emph{cow} detection to an
    incorrect \emph{horse}; TaskGuard preserves the input.
    \textbf{(c)} On natural rain, deraining introduces a false-positive
    \emph{car}; TaskGuard preserves the input.
    Dashed boxes denote ground truth and solid boxes detector predictions.
    }
    \label{fig:qualitative_utility}
\end{figure*}

For instance, Figure~\ref{fig:qualitative_utility}(a) shows a clear benefit
from restoration: the detector correctly detects all ground-truth objects in
the restored image, while missing all of them in the degraded counterpart.
However, the effect of restoration can be surprisingly subtle.
Figure~\ref{fig:qualitative_utility}(b) shows a restored image that appears
cleaner and remains visually close to its degraded input.
Nevertheless, the degraded image yields the correct \emph{cow} detection,
whereas restoration leads to a higher-confidence but incorrect \emph{horse}
prediction.
Similarly, in Figure~\ref{fig:qualitative_utility}(c), restoration introduces
a false-positive \emph{car} detection in the lower-right corner, even though
no car is present.
Thus, perceptual clarity and detector confidence can improve even when detector
correctness deteriorates.
This motivates evaluating the effect of the \emph{specific restoration
intervention} on the downstream task rather than relying on visual quality or
detector confidence alone.

We therefore consider a post-hoc decision problem:
\emph{given a degraded image and the particular restoration produced by a
frozen restorer, should that candidate restoration actually be accepted?}
We call this \emph{restoration utility prediction}.
Unlike approaches that modify the enhancement process using downstream
supervision~\cite{liu2022iayolo,kalwar2023gdip,kim2024sr4ir,chen2025unirestore},
we leave both the restorer and detector unchanged. Instead, TaskGuard estimates
the task utility of the realized image change and decides whether to restore or
preserve the degraded input.
Accordingly, it selects \emph{restore} for
Figure~\ref{fig:qualitative_utility}(a) and \emph{preserve} for (b) and (c).

The key hypothesis is that restoration utility depends not only on what changed
but on whether that change agrees with downstream task sensitivity.
TaskGuard therefore couples the realized restoration residual with a detector
gradient.
We validate this directional mechanism using exact regional counterfactual
interventions and additionally test whether its annotation-free pseudo-gradient
counterpart preserves the privileged signal used in the analysis.

The TaskGuard utility predictor is trained only on Gaussian degradation and
frozen before evaluation on fresh Gaussian data, unseen motion blur, rain, and
defocus, and natural-rain DAWN~\cite{kenk2020dawn}.
The same frozen decisions are also evaluated under
RT-DETR-L~\cite{zhao2024rtdetr}.
Our contributions are:

\begin{itemize}
    \item We formulate restoration utility as a post-hoc intervention for
    frozen restoration--detection pipelines and introduce a directional
    residual--task-sensitivity representation.

    \item We validate the proposed mechanism with exact regional
    counterfactuals, direct pseudo-gradient analysis, and feature-group
    ablation showing that task-conditioned evidence adds information beyond
    detector-response and residual statistics.

    \item We develop a frozen Restore-or-Preserve policy with a TRAIN-only
    risk--utility threshold and demonstrate zero-shot degradation-family
    transfer, natural-rain transfer, practical-risk reduction, and partial
    cross-detector safety transfer.
\end{itemize}

\section{Related Work}
\label{sec:related_work}
\paragraph{Task-aware enhancement and restoration.}
Detection-oriented enhancement methods couple image processing with downstream
objectives.
IA-YOLO~\cite{liu2022iayolo} predicts differentiable processing parameters,
GDIP~\cite{kalwar2023gdip} learns gated combinations of processing operators,
and Detection-Friendly Dehazing~\cite{li2023detectionfriendly} explicitly
couples dehazing with detection.
Task-driven restoration further uses recognition supervision to shape the
restoration itself, including SR4IR~\cite{kim2024sr4ir},
UniRestore~\cite{chen2025unirestore},
EDTR~\cite{kim2025edtr}, and TaskTok~\cite{lee2026tasktok}.
TaskGuard instead assumes that restoration and detection are frozen and
controls whether an already-generated candidate is accepted.

\paragraph{Machine-oriented quality and enhancement selection.}
Machine-oriented image assessment shows that perceptual quality does not
necessarily reflect downstream model preference
~\cite{li2025machinepreference}.
Most closely related, LQTARanker~\cite{fang2026lqtaranker} predicts
detection-performance gain and makes a per-image enhance-or-not decision.
TaskGuard instead conditions utility prediction on the interaction between the
\emph{actual restoration residual} and detector sensitivity, and validates this
directional mechanism against exact regional counterfactual interventions.
In addition, the TaskGuard utility predictor is trained only on Gaussian
degradation and evaluated across unseen degradation families, with the decision
threshold frozen before final testing.
Our focus is therefore not enhancement selection itself, but
restoration-specific directional task utility for post-hoc risk control.

\section{Method}
\label{sec:method}

\begin{figure*}[t]
    \centering
    \includegraphics[width=0.92\textwidth]
    {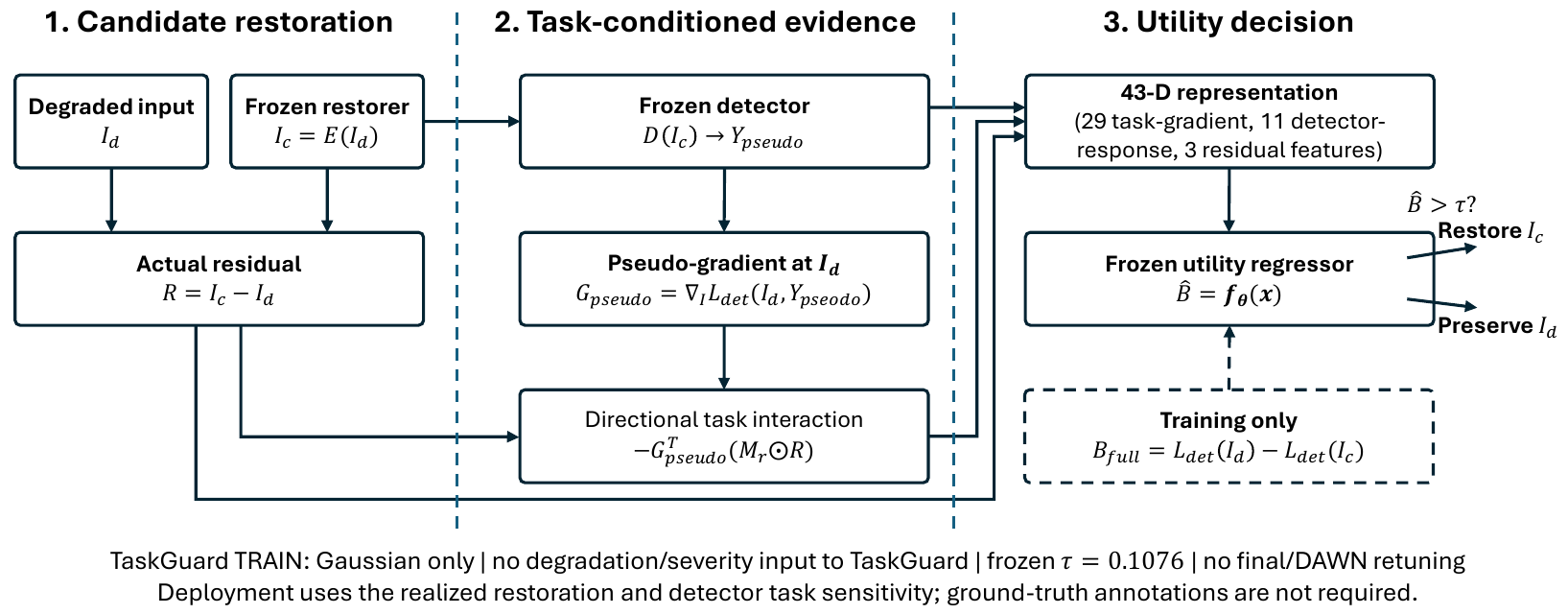}
    \caption{
    TaskGuard overview.
    A frozen restorer generates $I_c$; TaskGuard summarizes task-conditioned,
    detector-response, and residual evidence to predict utility and restores
    only when $\hat B>\tau$.
    }
    \label{fig:taskguard_overview}
\end{figure*}

Figure~\ref{fig:taskguard_overview} summarizes the TaskGuard pipeline.
A frozen restorer first produces a candidate restoration, after which
TaskGuard characterizes the realized image change using task-conditioned,
detector-response, and residual evidence to predict if the candidate
should be accepted. Let a frozen restorer $E$ produce candidate $I_c$ from degraded input $I_d$:
\begin{equation}
    I_c=E(I_d), \qquad R=I_c-I_d ,
    \label{eq:residual}
\end{equation}
where $R$ is the realized restoration intervention.

\subsection{Restoration Utility}

With detector loss $L_{\mathrm{det}}(I;Y)$, exact whole-image utility is
\begin{equation}
    B_{\mathrm{full}}
    =
    -\left (L_{\mathrm{det}}(I_c;Y)
    -
    L_{\mathrm{det}}(I_d;Y)\right ).
    \label{eq:bfull}
\end{equation}
Positive values favor restoration; $B_{\mathrm{full}}<0$ denotes a
\emph{loss-negative} intervention.
Detector loss is used as a differentiable per-image utility signal and is not
treated as equivalent to AP or per-image F1.

For spatial mask $M_r$, the exact regional counterfactual and its utility are
\begin{equation}
\begin{aligned}
    I_r^{\mathrm{CF}} &= I_d+M_r\odot R,\\
    B_r^{\mathrm{CF}}
    &=L_{\mathrm{det}}(I_d;Y)
      -L_{\mathrm{det}}(I_r^{\mathrm{CF}};Y).
\end{aligned}
\label{eq:regional_cf}
\end{equation}
A first-order (Taylor) directional estimate is
\begin{equation}
    B_r^{\mathrm{grad}}
    =
    -
    \nabla_{I_d}L_{\mathrm{det}}(I_d;Y)^\top
    (M_r\odot R).
    \label{eq:regional_grad}
\end{equation}
The residual specifies the actual image-change direction; the gradient
specifies local task sensitivity.
Equation~\eqref{eq:regional_grad} is directional evidence rather than an exact
counterfactual estimator.

\subsection{Deployable Task Signal}

Ground truth is unavailable at deployment.
TaskGuard obtains pseudo-labels ($Y_p$) from the restored candidate and evaluates the
corresponding gradient ($G_p$) at the degraded input:
\begin{equation}
    Y_p=\mathcal{P}_{0.25}(D(I_c)),
    \qquad
    G_p=\nabla_{I_d}L_{\mathrm{det}}(I_d;Y_p).
    \label{eq:pseudograd}
\end{equation}
The deployable regional interaction becomes
\begin{equation}
    B_r^{p}=-G_p^\top(M_r\odot R).
    \label{eq:pseudo_benefit}
\end{equation}

\subsection{Intervention Representation}

TaskGuard uses a $43$-D representation
\begin{equation}
    x=[x_{\mathrm{task}},x_{\mathrm{det}},x_{\mathrm{res}}],
    \qquad
    x\in\mathbb{R}^{29+11+3}.
    \label{eq:features}
\end{equation}
The $29$ task features summarize $64$ regional scores $\{B_r^p\}$ through
distribution statistics, positive/negative score mass, signed top-$k$
statistics for $k\in\{1,2,4,8,16,32\}$, and pseudo-supervised detector loss.
The $11$ detector-response features contain detection count and confidence
statistics for $I_d$ and $I_c$ together with their differences.
The $3$ residual features are the mean, standard deviation, and maximum of the elementwise absolute residual $|R|$: $\operatorname{mean}(|R|)$, $\operatorname{std}(|R|)$, and $\max(|R|)$.
No degradation family, severity, or restorer identity is provided to the
TaskGuard utility predictor.
Exact feature definitions are given in the supplementary material.

\subsection{Utility Regression and Frozen Decision}

A standardized Ridge regressor ($\alpha=10$) predicts restoration utility; its $\ell_2$ regularization provides a low-capacity fit to the correlated $43$-D summary features while limiting overfitting to the training distribution.
Each regressor is trained on the Gaussian TRAIN set to predict the exact whole-image utility target $B_{\mathrm{full}}$.
Ten repeated five-fold source-grouped stratified splits produce $50$ fold-specific regressors, which constitute the frozen inference ensemble.
For each candidate pair, their predictions are averaged to obtain $\hat B$. The policy is
\begin{equation}
    \pi(I_d,I_c)=
    \begin{cases}
        I_c,&\hat B>\tau,\\
        I_d,&\hat B\leq\tau.
    \end{cases}
    \label{eq:policy}
\end{equation}

The threshold is selected from \emph{Gaussian TRAIN out-of-fold predictions
only}.
Among thresholds retaining at least $98\%$ of Always-Restore mean utility, we
choose the one with the lowest loss-negative rate, breaking ties by higher
retained utility and then lower preserve rate.
This procedure yields $\tau=0.1076$.
Validation and final-benchmark data are not used to select or retune the
threshold.

\section{Experimental Setup}
\label{sec:experiments}

\paragraph{Controlled data.}
All controlled source images come from COCO 2017
(\texttt{val2017})~\cite{lin2014coco}.
The development pool contains $500$ sources with Gaussian noise
$\sigma\in\{15,25,50\}$, split by source into
$350/50/100$ TRAIN/VAL/development-TEST images
($1050/150/300$ pairs).

After the complete policy is frozen, we evaluate on a separate,
source-disjoint set of $200$ untouched COCO \texttt{val2017} images.
Each source is evaluated under Gaussian noise, motion blur, synthetic rain, and
defocus at three severities, yielding $2400$ pairs.

\paragraph{Models and natural data.}
Candidate restorations use frozen pretrained Restormer
specialists~\cite{zamir2022restormer}, with the corresponding specialist used
for each controlled degradation family.
TaskGuard receives only the resulting degraded/candidate pair and is not given
the degradation family, severity, or restorer identity.
The primary detector is COCO-pretrained YOLO11s
~\cite{yolo11_ultralytics} at input size $640$.
We additionally evaluate frozen TaskGuard decisions with
RT-DETR-L~\cite{zhao2024rtdetr} on the $600$ fresh Gaussian pairs.
Natural-domain transfer uses $198$ rainy images from
DAWN~\cite{kenk2020dawn}; no DAWN data are used for fitting or tuning. For DAWN, the person, bicycle, car, motorcycle, bus, and truck annotations are mapped directly to the corresponding COCO categories, and COCO-style AP$_{50:95}$ is computed over the 198 rainy images used in our evaluation.

\paragraph{Baselines.}
We compare TaskGuard against both fixed policies and learned appearance-based
utility predictors.
Always Preserve uses $I_d$ for every sample and represents the no-restoration
reference, while Always Restore uses $I_c$ unconditionally and therefore
measures the aggregate benefit and risk of applying every candidate restoration.
The learned baselines test whether restoration utility can be predicted from
generic visual appearance without the proposed task-conditioned directional
signal.
Image-only uses frozen ResNet-50 features~\cite{he2016resnet} from $I_d$ and
asks whether the degraded observation alone is sufficient to predict whether
restoration will help.
The stronger Visual-pair baseline observes both images through
\[
[f_d,f_c,f_c-f_d,|f_c-f_d|],
\]
and can therefore capture the appearance of the degraded and restored images
as well as the magnitude and direction of their feature-space change.
However, neither appearance baseline explicitly measures whether the realized
restoration change agrees with detector sensitivity.
Thus, comparison with Visual-pair is particularly important for isolating the
value of TaskGuard's task-conditioned gradient--residual interaction beyond
simply observing both images.
All learned baselines follow the same Gaussian-only development and freezing
protocol, with no fitting or retuning on the unseen degradation families.

\paragraph{Metrics.}
The reported metrics capture complementary aspects of utility prediction,
intervention risk, and downstream detection performance.
\textbf{Spearman correlation} $\rho_s(\hat B,B_{\mathrm{full}})$ measures whether the
predicted utility correctly ranks restorations from less to more beneficial,
without requiring the predicted and exact utilities to have the same numerical
scale.
\textbf{Benefit retention},$\frac{\mathbb{E}[B_\pi]}{\mathbb{E}[B_{\mathrm{full}}]}$,
measures how much of the mean utility of Always Restore is retained by the
selective policy; values above $100\%$ are possible when preserving
negative-utility candidates increases mean policy utility.
\textbf{Loss-negative reduction} measures the relative decrease, compared with Always
Restore, in interventions for which $B_{\mathrm{full}}<0$, and therefore
directly quantifies avoidance of restoration decisions that worsen detector
loss.
\textbf{Preserve rate} reports how often the policy rejects the restoration and retains
$I_d$, indicating how conservative the policy is.
We additionally report standard \textbf{COCO bounding-box} $\mathrm{AP}_{50:95}$ to
verify that reducing negative interventions does not come at an excessive cost
in aggregate detection accuracy.

Because detector loss is only a differentiable proxy for task quality, we also
evaluate an operational notion of deterioration.
At detector confidence $0.25$ and IoU $0.50$, let
$\Delta F1=F1_c-F1_d$ and $\Delta FN=FN_c-FN_d$.
A restoration is \emph{practical-negative} when $\Delta F1<0$ or
$\Delta FN>0$, i.e., when restoration decreases\textbf{ per-image F1} or increases the
number of \textbf{missed ground-truth objects}.
This provides a task-level complement to the loss-defined negative-intervention
measure.
All statistical resampling is performed at the source-image level so that
degradation variants and regional observations derived from the same clean
source remain grouped rather than being treated as independent samples.
\section{Results}
\label{sec:results}

\subsection{Directional Utility Approximation}

\begin{figure*}[!htb]
    \centering
    \includegraphics[width=\linewidth]
    {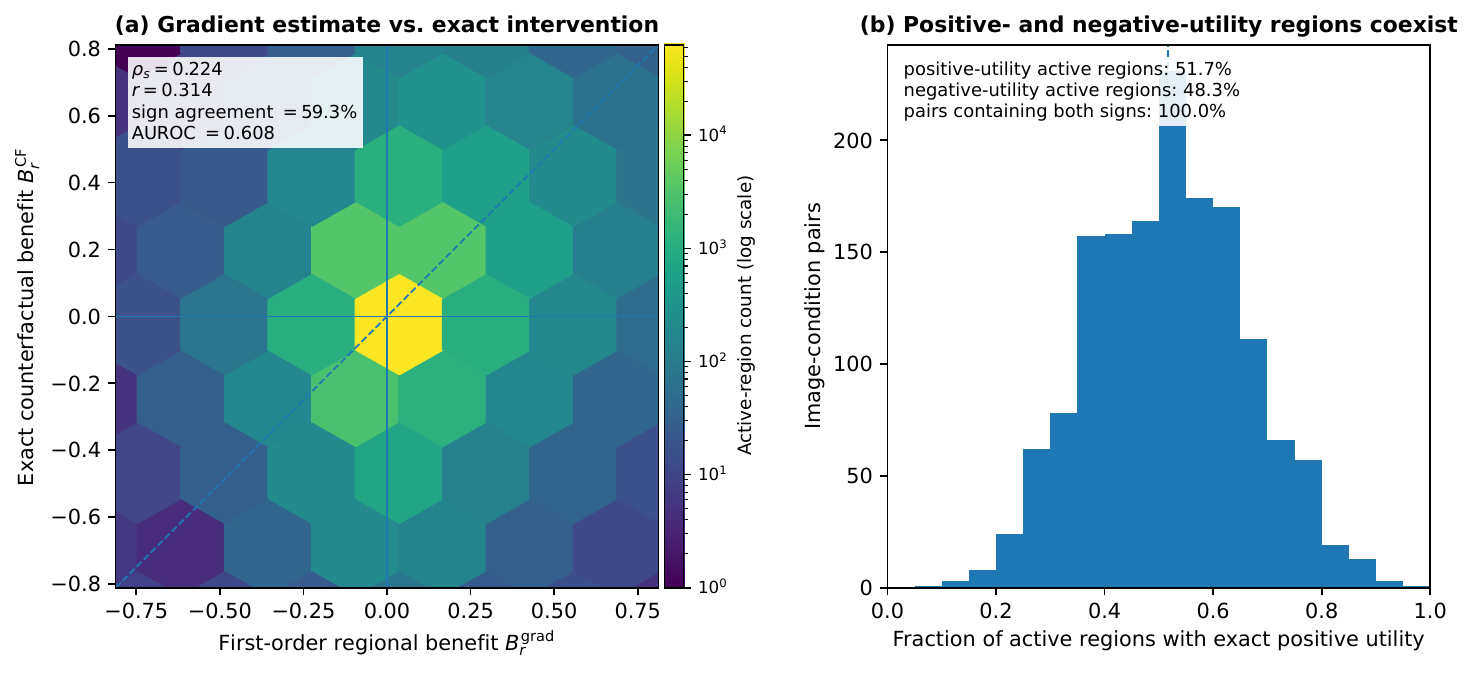}
    \caption{
    Regional restoration utility and its directional approximation.
    Helpful and harmful regions coexist within every image--condition pair,
    while the first-order signal is imperfect but systematically associated
    with exact counterfactual utility.
    }
    \label{fig:counterfactual_mechanism}
\end{figure*}

\paragraph{Regional utility heterogeneity.}
Across the full $500$-source Gaussian development pool, comprising
$96{,}000$ regional interventions, $83{,}088$ regions contain an active
restoration change.
Among active regions, $51.7\%$ have positive utility and $48.3\%$
negative utility, and every image--condition pair contains both signs.
Thus, whole-image restoration hides substantial within-image heterogeneity.

\paragraph{Privileged directional estimate.}
The privileged directional estimate in Eq.~\eqref{eq:regional_grad} is
imperfect but informative:
Pearson $r=0.314$, Spearman $\rho_s=0.224$, sign agreement $59.3\%$, and
AUROC $0.608$ against exact regional counterfactual utility.
Figure~\ref{fig:counterfactual_mechanism} visualizes both the pronounced
within-image utility heterogeneity and the systematic, though imperfect,
relationship between directional and exact counterfactual utility.

\paragraph{Pseudo-gradient validation.}
On the Gaussian TRAIN subset, comprising $58{,}512$ active regions from
$350$ sources and $1050$ degraded/restored pairs,
$B_r^p$ remains positively associated with exact $B_r^{\mathrm{CF}}$:
Pearson $r=0.261$ (95\% CI $[0.213,0.309]$),
Spearman $\rho_s=0.144$ ($[0.129,0.159]$), and
AUROC $0.566$ ($[0.559,0.572]$).
It is substantially more aligned with the privileged directional signal
$B_r^{\mathrm{grad}}$:
$r=0.767$ ($[0.680,0.834]$),
$\rho_s=0.554$ ($[0.526,0.581]$), and
$71.8\%$ sign agreement ($[70.5,73.0]\%$).
Thus, pseudo-labeling attenuates but does not remove the proposed directional
task information, providing development-time evidence that the deployable
pseudo-gradient preserves the intended mechanism and remains meaningfully
related to exact counterfactual utility.
\subsection{Fresh Controlled Evaluation}

\begin{table*}[t]
\centering
\caption{
Fresh controlled evaluation of the frozen TaskGuard policy.
Gaussian is the training degradation; motion, rain, and defocus are evaluated
zero-shot.
}
\label{tab:controlled_final}

\begin{threeparttable}
\small

\begin{tabular*}{\textwidth}{@{\extracolsep{\fill}}lrrrrrrrr@{}}
\toprule
Family & $B_{\mathrm{restore}}$ & Benefit Ret. & Neg. Restore &
Neg. TG & Neg. Red. & AP$_{\mathrm{restore}}$ &
AP$_{\mathrm{TG}}$ & AP Ret. \\
\midrule
Gaussian & 1.063 & 97.0\% & 19.5\% & 11.0\% & 43.6\% &
0.5165 & 0.5096 & 98.7\% \\
\midrule
Motion & 2.003 & 97.6\% & 7.8\% & 4.8\% & 38.3\% &
0.5059 & 0.5013 & 99.1\% \\
Rain & 0.365 & 89.1\% & 32.5\% & 7.7\% & 76.4\% &
0.5456 & 0.5354 & 98.1\% \\
Defocus & 0.460 & 93.1\% & 24.7\% & 12.8\% & 48.0\% &
0.4863 & 0.4832 & 99.4\% \\
\textbf{Zero-shot macro} &
\textbf{0.942} & \textbf{93.3\%} & \textbf{21.7\%} &
\textbf{8.4\%} & \textbf{54.2\%} &
\textbf{0.5126} & \textbf{0.5066} & \textbf{98.8\%} \\
\bottomrule
\end{tabular*}

\begin{tablenotes}[flushleft]
\footnotesize
\item[]
$B_{\mathrm{restore}}$: mean detector-loss benefit of Always Restore;
Ret.: retention; Neg.: $B_{\mathrm{full}}<0$; Red.: reduction;
TG: TaskGuard; AP Ret.: TaskGuard AP$_{50:95}$ divided by Always-Restore
AP$_{50:95}$.
\end{tablenotes}

\end{threeparttable}
\end{table*}

Table~\ref{tab:controlled_final} summarizes the performance of the frozen
TaskGuard policy across the fresh controlled degradation families. On Gaussian noise, TaskGuard retains $97.0\%$ of loss-defined restoration
benefit while reducing the loss-negative rate from $19.5\%$ to $11.0\%$.
Without family-specific retuning, the same policy reduces loss-negative
interventions by $38.3\%$, $76.4\%$, and $48.0\%$ for motion blur, rain, and
defocus, respectively.

Across the three unseen families, TaskGuard retains $93.3\%$ of loss-defined
benefit and $98.8\%$ of Always-Restore COCO AP while reducing loss-negative
interventions by $54.2\%$.
Always Restore remains best in aggregate AP; TaskGuard therefore controls
intervention risk rather than maximizing AP.

\subsection{What Information Drives Utility Prediction?}

\begin{table*}[t]
\centering
\begin{threeparttable}

\caption{
Representation comparison on unseen motion, rain, and defocus, with paired
TaskGuard gains over Visual-pair.
}
\label{tab:representation}

\small

\begin{tabular}{@{}
    p{0.61\textwidth}
    @{\hspace{0.03\textwidth}}
    p{0.36\textwidth}
    @{}}

\centering\textbf{A. Representation comparison}
&
\centering\textbf{B. TaskGuard gain over Visual-pair}
\tabularnewline[1mm]

\begin{minipage}[t]{\linewidth}
\vspace{0pt}
\centering
\setlength{\tabcolsep}{3.4pt}

\begin{tabular}{lrrrrr}
\toprule
Representation & $\rho_s$ & Benefit Ret. & Neg. Red. &
Prac. Red. & Preserve \\
\midrule
Image-only
& 0.091 & 86.1\% & 6.9\% & 12.4\% & 9.3\% \\

Visual-pair
& 0.211 & 79.6\% & 30.3\% & 21.8\% & 23.5\% \\

\textbf{TaskGuard}
& \textbf{0.622} & \textbf{93.3\%} &
\textbf{54.2\%} & \textbf{37.0\%} & 32.7\% \\
\bottomrule
\end{tabular}
\end{minipage}

&

\begin{minipage}[t]{\linewidth}
\vspace{0pt}
\centering
\setlength{\tabcolsep}{4.0pt}

\begin{tabular}{lrr}
\toprule
Metric & Gain & 95\% CI \\
\midrule
Benefit ret.
& +19.50 pp & [+14.22, +25.13] \\

Neg. red.
& +33.59 pp & [+25.57, +41.75] \\

Prac. red.
& +15.15 pp & [+7.71, +22.84] \\
\bottomrule
\end{tabular}
\end{minipage}

\tabularnewline
\end{tabular}
\vspace{3mm}
\begin{tablenotes}[flushleft]
\footnotesize
\item[]
$\rho_s$, benefit retention, loss-negative reduction, and preserve rate are
unweighted family macros; practical-negative reduction is pooled over all
1,800 zero-shot pairs. Part B uses paired source bootstrap.
Ret.: retention; Neg.: loss-negative; Prac.: practical-negative;
Red.: reduction; CI: confidence interval; pp: percentage points.
\end{tablenotes}

\end{threeparttable}
\end{table*}

Table~\ref{tab:representation} compares TaskGuard with the
Image-only and Visual-pair representations on the unseen degradation families. Image-only and Visual-pair predictors achieve zero-shot
$\rho_s=0.091$ and $0.211$, respectively, versus $0.622$ for TaskGuard.
Relative to Visual-pair, TaskGuard improves benefit retention by $19.5$
percentage points (95\% CI $[14.2,25.1]$), loss-negative reduction by $33.6$
points ($[25.6,41.8]$), and practical-negative reduction by $15.2$ points
($[7.7,22.8]$).

A post-freeze feature-group ablation further isolates the proposed task
signal.
Task-gradient features alone achieve $\rho_s=0.631$, compared with $0.480$
for detector-response and $0.240$ for residual-only features.
Adding task-gradient features to Detector+Residual raises
$\rho_s$ from $0.470$ to $0.622$:
\[
    \Delta\rho_s=0.152,
    \qquad
    95\%\ \mathrm{CI}=[0.110,0.197],
\]
while increasing benefit retention by $6.6$ points
($[2.5,10.6]$).
Loss-negative reduction also increases by $6.9$ points, although its
95\% CI $[-0.2,14.3]$ includes zero.
The complete seven-group ablation is reported in the supplement.

\subsection{Frozen Threshold and Risk--Utility Tradeoff}

\begin{figure*}[t]
    \centering
    \includegraphics[width=\textwidth]
    {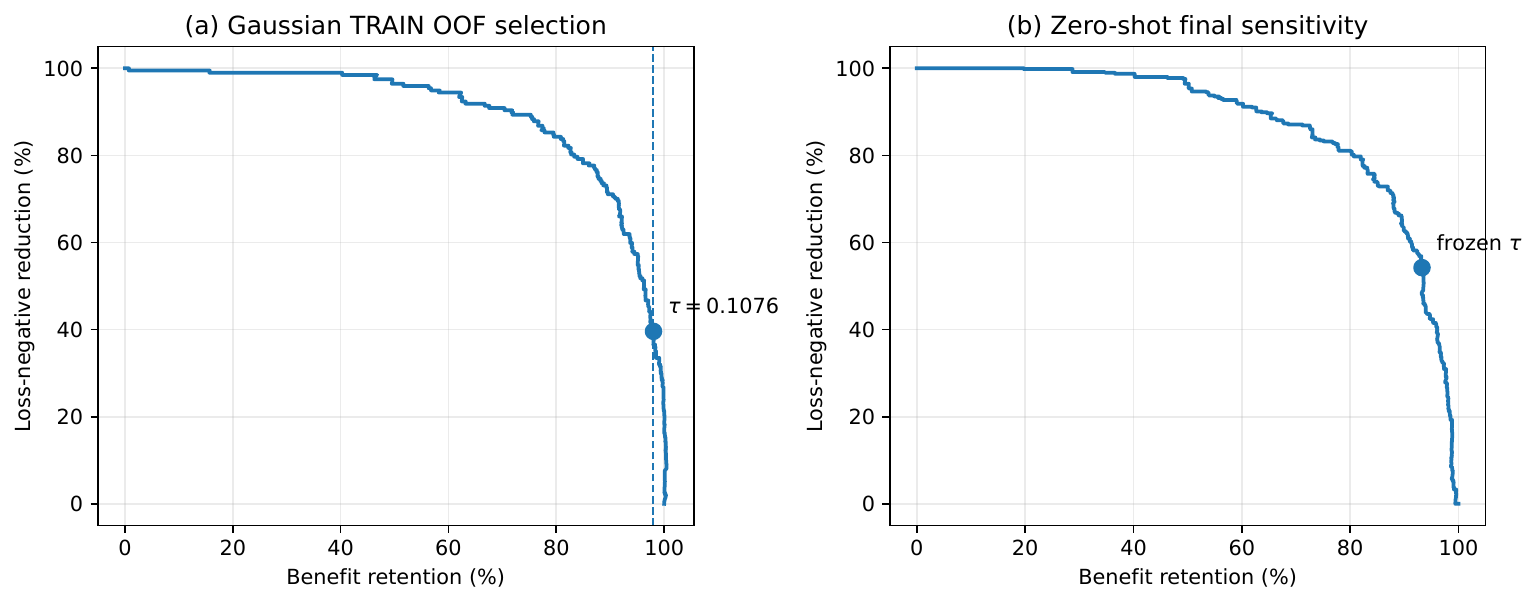}
    \caption{
    Risk--utility threshold analysis.
    \textbf{(a)} Gaussian TRAIN OOF selection curve with the predefined $98\%$
    benefit-retention constraint.
    \textbf{(b)} Post-freeze sensitivity on unseen degradations; this panel is
    descriptive only.
    }
    \label{fig:threshold_risk_utility}
\end{figure*}

The operating threshold is determined entirely from Gaussian TRAIN OOF
predictions.
The predefined $98\%$-retention rule yields $\tau=0.1076$, retaining
$98.0\%$ of TRAIN Always-Restore utility while reducing loss-negative
interventions by $39.6\%$.
Figure~\ref{fig:threshold_risk_utility} shows the TRAIN OOF curve used for
this selection together with the post-freeze sensitivity of the same operating
points on the unseen degradation families.

Nearby TRAIN-selected operating points exhibit a smooth tradeoff:
retention constraints of $99\%$, $98\%$, and $97\%$ give TRAIN
loss-negative reductions of $33.5\%$, $39.6\%$, and $46.7\%$.
Applied unchanged to the unseen families, the same thresholds give
benefit-retention/risk-reduction pairs of
$93.9/45.5\%$, $93.3/54.2\%$, and $91.0/60.3\%$, respectively.
The frozen $98\%$-retention operating point is therefore one reproducible point
along the observed risk--utility tradeoff rather than a test-selected
threshold.

\subsection{Practical and Cross-Detector Risk}

\begin{table*}[t]
\centering

\caption{
Cross-detector transfer and operational evaluation of the frozen TaskGuard
policy.
}
\label{tab:cross_detector_practical}

\begin{threeparttable}
\small
\begin{tabular*}{\textwidth}{
@{\extracolsep{\fill}}
lrrrrrr
@{}
}
\toprule
\multicolumn{7}{l}{\textbf{A. Cross-detector transfer on fresh Gaussian data}} \\
Utility evaluator & $\rho_s$ & Benefit Ret. & Neg. Restore &
Neg. TG & Neg. Red. & Preserve \\
\midrule
YOLO11s
& 0.726 & 97.0\% & 19.5\% & 11.0\% & 43.6\% & 22.2\% \\
\textbf{RT-DETR-L}
& 0.305 & 93.9\% & 28.5\% & 19.8\% & 30.4\% & 22.2\% \\
\midrule

\multicolumn{7}{l}{\textbf{B. Practical meaning of loss-defined utility at confidence 0.25}} \\
Scope & Neg. Rate & Prac. Rate &
$P(\mathrm{Prac.}\mid\mathrm{Neg.})$ &
$P(\mathrm{Prac.}\mid\neg\mathrm{Neg.})$ &
$\rho_s(B_{\mathrm{full}},\Delta F1)$ &
Prac. Red. \\
\midrule
All four families
& 21.1\% & 18.1\% & 34.3\% & 13.7\% & 0.601 & 34.1\% \\
\textbf{Zero-shot pooled}
& 21.7\% & 18.3\% & 32.3\% & 14.5\% & 0.587 & 37.0\% \\
\bottomrule
\end{tabular*}

\begin{tablenotes}[flushleft]
\footnotesize
\item[]
Zero-shot denotes pooled motion, rain, and defocus pairs.
Practical-negative means per-image F1 decreases or missed ground-truth objects
increase at confidence $0.25$ and IoU $0.50$.
Ret.: retention; Neg.: loss-negative; Prac.: practical-negative;
Red.: reduction; TG: TaskGuard.
\end{tablenotes}

\end{threeparttable}
\end{table*}

Table~\ref{tab:cross_detector_practical} summarizes cross-detector transfer
and the practical interpretation of loss-defined restoration utility. On the fresh Gaussian set, frozen TaskGuard decisions retain $93.9\%$ of
RT-DETR-L loss-defined restoration benefit while reducing RT-DETR-L
loss-negative interventions from $28.5\%$ to $19.8\%$.
Utility-score correlation is lower under RT-DETR-L
($\rho_s=0.305$) than YOLO11s ($0.726$), so we characterize this as
\emph{partial cross-detector safety transfer}, not detector invariance.

Across all controlled families, $21.1\%$ of restorations are loss-negative and
$18.1\%$ practical-negative.
Loss-defined utility correlates with $\Delta F1$ at $\rho_s=0.601$, and
$-B_{\mathrm{full}}$ attains AUROC $0.700$ for identifying practical-negative
restorations.
Across the three unseen families, TaskGuard reduces the practical-negative
rate from $18.3\%$ to $11.6\%$, a $37.0\%$ relative reduction.

\subsection{Natural-Rain Transfer}

\begin{table*}[t]
\centering
\caption{
Frozen-policy transfer to 198 natural-rain DAWN images.
}
\label{tab:dawn}

\begin{threeparttable}
\small

\begin{tabular*}{\textwidth}{
@{\extracolsep{\fill}}
lrrrrrr
@{}
}
\toprule

\multicolumn{7}{l}{\textbf{A. Frozen-policy performance on natural-rain DAWN}} \\
Policy & \# Restored & AP$_{50:95}$ & AP Gain Ret. &
Benefit Ret. & Neg. Red. & Prac. Red. \\
\midrule

Original
& 0 & 0.3578 & 0.0\% & -- & -- & -- \\

Always Derain
& 198 & 0.3818 & 100.0\% & 100.0\% & 0.0\% & 0.0\% \\

Image-only
& 189 & 0.3805 & 94.5\% & 99.7\% & 6.4\% & 16.7\% \\

Visual-pair
& 41 & 0.3773 & 81.3\% & 101.2\% & 80.9\% & 83.3\% \\

\textbf{TaskGuard}
& 8 & 0.3765 & 77.8\% & 100.4\% &
\textbf{97.9\%} & \textbf{100.0\%} \\

\midrule

\multicolumn{7}{l}{\textbf{B. Paired source-bootstrap comparisons}} \\
\multicolumn{4}{l}{Comparison}
& \multicolumn{1}{r}{Estimate}
& \multicolumn{2}{r}{95\% CI} \\
\midrule

\multicolumn{4}{l}{TaskGuard $-$ Visual-pair AP$_{50:95}$}
& -0.0008
& \multicolumn{2}{r}{[-0.0028, +0.0010]} \\

\multicolumn{4}{l}{TaskGuard $-$ Always Derain AP$_{50:95}$}
& -0.0053
& \multicolumn{2}{r}{[-0.0133, +0.0021]} \\

\multicolumn{4}{l}{Neg. Red. gain vs.\ Visual-pair}
& +17.0 pp
& \multicolumn{2}{r}{[+8.4, +25.8]} \\

\bottomrule
\end{tabular*}

\begin{tablenotes}[flushleft]
\footnotesize
\item[]
AP Gain Ret.: fraction of the Always-Derain AP$_{50:95}$ improvement over
Original retained; Benefit Ret.: detector-loss utility retention.
Prac.: practical-negative (per-image F1 decrease or additional false
negatives); only six Always-Derain practical-negative cases occur, so
Prac. Red. is descriptive.
Neg.: loss-negative; Red.: reduction; AP: average precision;
pp: percentage points; CI: confidence interval.
\end{tablenotes}

\end{threeparttable}
\end{table*}

Table~\ref{tab:dawn} summarizes the natural-rain DAWN evaluation
and comparison with the frozen appearance-based baselines. On $198$ natural-rain DAWN images, original observations achieve COCO-style AP
$0.3578$ and unconditional deraining reaches $0.3818$.
TaskGuard restores only $8$ images and obtains AP $0.3765$, retaining $77.8\%$
of the Always-Derain AP gain while reducing loss-negative interventions by
$97.9\%$.
Benefit retention can exceed $100\%$ because preserving selected
negative-utility candidates can yield greater mean policy utility than
unconditional deraining.

Visual-pair retains $81.3\%$ of the AP gain but reduces loss-negative
interventions by only $80.9\%$.
TaskGuard improves the latter by $17.0$ percentage points
(95\% CI $[8.4,25.8]$).
The paired TaskGuard--Always-Derain AP difference is $-0.0053$
(95\% CI $[-0.0133,0.0021]$); because the interval includes zero, we make no
equivalence or non-inferiority claim.
Only six Always-Derain cases are practical-negative, so DAWN practical-risk
results are descriptive.

\paragraph{Runtime.}
Runtime was measured on a single NVIDIA RTX A4000 GPU using YOLO11s at
input size $640$.
On $50$ Gaussian validation sources at $\sigma=25$, with $5$ warm-up
samples, detector-only inference requires $26.7$ ms/image.
Once a restoration candidate is available, TaskGuard adds $136.1$ ms/image,
for a complete post-restoration stage of $162.9$ ms/image.
Using the separately measured Restormer directory-run timing, the contextual
Restore+Detector and Restore+TaskGuard pipelines require approximately
$667$ and $803$ ms/image, respectively.

\section{Discussion}
\label{sec:discussion}

\paragraph{Risk control rather than AP maximization.}
TaskGuard deliberately trades a small amount of aggregate AP for fewer
negative interventions.
Its threshold is fixed using a TRAIN-only utility-retention constraint rather
than final performance, and applications with different intervention costs can
choose another operating point using development data.

\paragraph{Directional evidence is not an oracle.}
Both the privileged first-order approximation and deployable pseudo-gradient
are imperfect.
The pseudo-gradient has only modest direct agreement with exact regional
counterfactual utility, but its source-bootstrap intervals remain above the
corresponding null or chance baselines, and the pseudo signal remains
substantially aligned with its privileged counterpart.
Feature-group ablation independently shows that aggregating this
task-conditioned evidence improves zero-shot whole-image utility prediction.
Because deployment pseudo-labels are obtained from $I_c$, confident
restoration-induced prediction errors can also corrupt the pseudo-gradient.
Uncertainty-aware pseudo-labeling or teacher ensembles are natural extensions.

\paragraph{Scope and computational cost.}
TaskGuard must generate $I_c$ before judging it and therefore does not avoid
restoration cost.
The gradient computation also adds nontrivial overhead.
Our zero-shot claim concerns degradation-family transfer from Gaussian training
to motion, rain, and defocus; DAWN provides an additional natural-domain test.
The experiments do not establish arbitrary transfer across restorers,
detectors, datasets, or tasks.

\paragraph{Loss-defined versus practical utility.}
Detector loss provides the differentiable per-image signal needed for training,
but it is not equivalent to AP or per-image F1.
The practical-negative analysis demonstrates a meaningful association between
loss-defined utility and concrete detection deterioration while also showing
that the relationship is not one-to-one.
Both levels of evaluation are therefore necessary.

\section{Conclusion}

TaskGuard treats restoration as a task-conditioned intervention rather than an
automatically beneficial preprocessing step.
It couples the actual restoration residual with frozen-detector sensitivity
and uses the resulting evidence to decide whether a candidate should be
accepted.
Exact regional counterfactuals, pseudo-gradient validation, and feature-group
ablation support the proposed directional mechanism as informative, although
not exact.
A TaskGuard policy whose utility predictor is trained only on Gaussian
degradation and frozen under a TRAIN-only risk--utility criterion transfers to
unseen controlled degradations and natural rain while substantially reducing
negative interventions and retaining most of the benefit of unconditional
restoration.
These results motivate restoration systems that explicitly account for the
downstream consequences of the particular image changes they introduce.

{
    \small
    \bibliographystyle{ieeenat_fullname}
    \bibliography{main}
}

\end{document}


\maketitle
\appendix

\section{Supplement Overview}

This supplement provides implementation details and analyses omitted from the
main paper for space:
(1) the exact $43$-D TaskGuard representation;
(2) the frozen $50$-model inference ensemble and threshold-selection protocol;
(3) the complete feature-group ablation;
(4) direct pseudo-gradient/counterfactual bootstrap analysis;
(5) threshold operating-point sensitivity;
and (6) additional representation, DAWN, cross-detector, practical-risk, and
runtime details.

\section{Exact TaskGuard Representation}
\label{supp:features}

TaskGuard represents each degraded/restored pair with
\[
x=[x_{\mathrm{task}},x_{\mathrm{det}},x_{\mathrm{res}}]
\in\mathbb{R}^{43},
\]
comprising $29$ task-gradient, $11$ detector-response, and $3$ residual
features.

The degraded and restored images are letterboxed identically to the detector
input resolution before the residual
\[
R=I_c-I_d
\]
is formed.
TaskGuard uses $64$ spatial masks arranged on an $8\times8$ grid.
The masks use the frozen feathering parameter \texttt{feather=8}; they are
therefore spatial masks rather than strictly binary blocks.

Pseudo-labels are obtained from restored-image detections at confidence
$0.25$.
The pseudo-gradient is evaluated at the degraded image,
\[
G_p=\nabla_{I_d}L_{\mathrm{det}}(I_d;Y_p),
\]
and the regional directional score is
\[
s_r
=
B_r^p
=
-
G_p^\top(M_r\odot R).
\]
Regional scores are sorted by descending \emph{signed} value, not absolute
magnitude.

Table~\ref{tab:supp_exact_features} lists the complete $43$-D
representation used by the frozen TaskGuard model.
\begin{table*}[t]
\centering
\small
\caption{Exact $43$-D TaskGuard feature representation.}
\label{tab:supp_exact_features}
\begin{tabularx}{\textwidth}{p{0.16\textwidth} p{0.08\textwidth} X}
\toprule
Group & Dim. & Features \\
\midrule
Task-gradient
& 29
&
\texttt{score\_sum},
\texttt{score\_mean},
\texttt{score\_std},
\texttt{score\_min},
\texttt{score\_max},
\texttt{score\_q10},
\texttt{score\_q25},
\texttt{score\_q50},
\texttt{score\_q75},
\texttt{score\_q90};
\texttt{positive\_count},
\texttt{positive\_fraction},
\texttt{positive\_mass},
\texttt{negative\_mass\_abs},
\texttt{positive\_minus\_negative\_mass},
\texttt{positive\_mass\_fraction};
for $k\in\{1,2,4,8,16,32\}$:
\texttt{score\_top$k$\_sum},
\texttt{score\_top$k$\_mean};
and \texttt{pseudo\_loss}.
\\
\midrule
Detector response
& 11
&
\texttt{degraded\_det\_count},
\texttt{degraded\_conf\_mean},
\texttt{degraded\_conf\_sum},
\texttt{degraded\_conf\_max},
\texttt{restored\_det\_count},
\texttt{restored\_conf\_mean},
\texttt{restored\_conf\_sum},
\texttt{restored\_conf\_max},
\texttt{det\_count\_delta},
\texttt{det\_conf\_sum\_delta},
\texttt{det\_conf\_mean\_delta}.
\\
\midrule
Residual
& 3
&
\texttt{residual\_abs\_mean},
\texttt{residual\_abs\_std},
\texttt{residual\_abs\_max}.
\\
\bottomrule
\end{tabularx}
\end{table*}

The task-gradient group can be written more compactly as follows.
For regional scores $\{s_r\}_{r=1}^{64}$, TaskGuard uses ten global
distribution statistics, six sign/mass statistics, and top-$k$ sum/mean pairs:
\[
T_k^{\mathrm{sum}}=\sum_{i=1}^{k}s_{(i)},\qquad
T_k^{\mathrm{mean}}=\frac{1}{k}\sum_{i=1}^{k}s_{(i)},
\]
where
$s_{(1)}\ge\cdots\ge s_{(64)}$
and
$k\in\{1,2,4,8,16,32\}$.
Together with the pseudo-supervised degraded-image loss, these yield $29$
task-gradient features.

No degradation family, severity, or restorer identity appears in the
representation.

\section{Frozen Training and Inference Protocol}
\label{supp:frozen_protocol}

The frozen candidate is
\texttt{all\_ridge\_reg}, implemented as
\texttt{StandardScaler + Ridge(alpha=10)}.
The Gaussian TRAIN set contains $350$ source images and $1050$ degraded/restored
pairs.

The inference predictor is \emph{not} a single refit on all TRAIN samples.
Instead, ten repeated five-fold source-grouped stratified splits produce
$50$ fold-specific regressors.
The stratification label indicates whether whole-image utility is negative,
and source image ID is the grouping variable.
For evaluation pair $x$, the frozen score is
\[
\hat B(x)
=
\frac{1}{50}
\sum_{m=1}^{50} f_m(x).
\]
These $50$ models constitute the frozen inference ensemble.

The implementation uses the frozen seed $2027$ and constructs the repeated-fold
random states from that record.
The final $2400$-pair locked benchmark table was generated by reconstructing
the frozen $50$-regressor ensemble from the recorded seed and fold schedule,
predicting each final pair with every regressor, and averaging their
predictions.
The final decision is
\[
\mathrm{Restore}
\iff
\hat B>0.107633802381.
\]

\subsection{TRAIN-only threshold selection}

The threshold is selected from Gaussian TRAIN out-of-fold scores only.
For threshold $t$, let
\[
\mathrm{Retain}(t)
=
\frac{\mathbb E[B_{\pi_t}]}
{\mathbb E[B_{\mathrm{full}}]}
\]
and let
$p^-(t)$ be the policy loss-negative intervention rate.
The frozen rule is
\[
\tau
=
\arg\min_t p^-(t)
\quad
\mathrm{s.t.}\quad
\mathrm{Retain}(t)\ge 0.98.
\]
Ties are resolved by higher mean benefit and then lower preserve rate.
The exact saved TRAIN curve reproduces
\[
\tau=0.107633802381
\]
to numerical precision.

Validation is used only as confirmation and does not retune the threshold.
At the frozen point, the validation policy has utility-ranking
$\rho_s=0.597$, benefit retention $95.9\%$, loss-negative rate $10.0\%$, and
relative loss-negative reduction $46.4\%$.

\section{Complete Feature-Group Ablation}
\label{supp:feature_ablation}

Table~\ref{tab:supp_feature_ablation} reports all seven feature-group variants
evaluated post-freeze on the three unseen controlled degradation families.
Each ablation follows the same Gaussian TRAIN protocol, and each threshold is
selected by the same TRAIN-only $98\%$-retention rule.

\begin{table*}[t]
\centering
\small
\caption{Complete zero-shot feature-group ablation on unseen controlled degradations.}
\label{tab:supp_feature_ablation}
\begin{threeparttable}
\begin{tabular*}{\textwidth}{@{\extracolsep{\fill}}lrrrrrrr@{}}
\toprule
Features & Dim. & $\tau$ & $\rho_s$ & Benefit ret. & Neg. red. & Preserve & Practical red. \\
\midrule
Residual
& 3 & 0.1549 & 0.240 & 0.882 & 0.195 & 0.154 & 0.209 \\
Detector
& 11 & 0.2998 & 0.480 & 0.896 & 0.433 & 0.275 & 0.309 \\
Detector+Residual
& 14 & 0.1823 & 0.470 & 0.866 & 0.474 & 0.299 & 0.348 \\
Task-gradient
& 29 & 0.0480 & \textbf{0.631} & \textbf{0.976} & 0.412 & 0.260 & 0.170 \\
Task-gradient+Residual
& 32 & 0.0250 & 0.632 & 0.974 & 0.427 & 0.269 & 0.206 \\
Task-gradient+Detector
& 40 & 0.1336 & 0.626 & 0.932 & \textbf{0.544} & 0.327 & 0.352 \\
All
& 43 & 0.1076 & 0.622 & 0.933 & 0.542 & 0.327 & \textbf{0.370} \\
\bottomrule
\end{tabular*}
\begin{tablenotes}[flushleft]
\footnotesize
\item[] Ret.: retention; Neg.: loss-negative; Practical: practical-negative.
Retention, reduction, and preserve-rate entries are fractions.
\end{tablenotes}
\end{threeparttable}
\end{table*}

Task-gradient features provide the strongest individual signal for utility
ranking.
The full representation is not claimed to maximize every metric:
Task-gradient+Residual attains slightly higher $\rho_s$, while
Task-gradient+Detector attains slightly higher loss-negative reduction.
The full $43$-D representation is the pre-frozen TaskGuard model used for all
primary evaluations.
Among the post-freeze ablations, it attains the highest practical-negative
reduction, while other subsets slightly exceed it on individual metrics.

\subsection{Paired source-bootstrap comparisons}

Table~\ref{tab:supp_feature_bootstrap} reports source-level paired bootstrap
comparisons for the two most informative contrasts.

\begin{table*}[t]
\centering
\small
\caption{Paired source-bootstrap differences for feature-group ablations.}
\label{tab:supp_feature_bootstrap}
\begin{threeparttable}
\begin{tabular}{llrr}
\toprule
Comparison & Metric & Difference & 95\% CI \\
\midrule
All $-$ Detector+Residual
& Spearman $\rho_s$ & +0.1519 & [0.1099, 0.1967] \\
& Benefit retention & +0.0661 & [0.0248, 0.1058] \\
& Loss-negative reduction & +0.0688 & [-0.0021, 0.1430] \\
& Practical-negative reduction & +0.0212 & [-0.0463, 0.0853] \\
& Preserve rate & +0.0278 & [0.0011, 0.0533] \\
\midrule
Task-gradient+Residual $-$ Residual
& Spearman $\rho_s$ & +0.3921 & [0.3312, 0.4491] \\
& Benefit retention & +0.0920 & [0.0603, 0.1296] \\
& Loss-negative reduction & +0.2319 & [0.1635, 0.2983] \\
& Practical-negative reduction & -0.0030 & [-0.0612, 0.0581] \\
& Preserve rate & +0.1156 & [0.0806, 0.1500] \\
\bottomrule
\end{tabular}
\begin{tablenotes}[flushleft]
\footnotesize
\item[] CI: source-bootstrap confidence interval.
\end{tablenotes}
\end{threeparttable}
\end{table*}

These comparisons support the conclusion that task-gradient features provide
information not captured by appearance-scale residual statistics or detector
response alone, while detector-response and residual features provide
complementary information for the final risk-aware policy.

\begin{figure}[t]
    \centering
    \includegraphics[width=\linewidth]
    {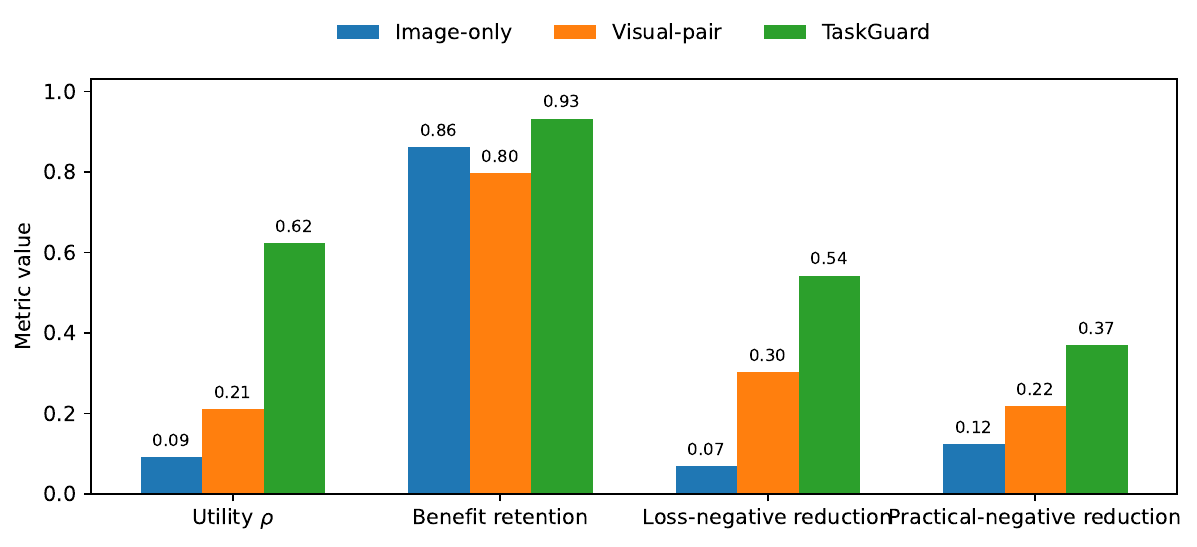}
    \caption{Representation comparison on unseen controlled degradation families.}
    \label{fig:supp_representation}
\end{figure}

Figure~\ref{fig:supp_representation} provides a complementary visualization
of the representation comparison reported quantitatively in the main paper.

\section{Pseudo-Gradient Versus Exact Counterfactual Utility}
\label{supp:pseudograd}

The direct pseudo-gradient analysis uses the cached Gaussian TRAIN regional
scores because $64$-region pseudo-gradient vectors were cached for the
$350$ TRAIN sources only.
This diagnostic is therefore explicitly a development-TRAIN mechanism analysis
and is not used for model or threshold selection.

The full $500$-source Gaussian development reference contains $96{,}000$
regional rows from $1500$ image--condition pairs.
A region is considered active when both the recorded residual and privileged
gradient activity are nonzero:
\[
\texttt{R\_abs\_mean}>0
\qquad\text{and}\qquad
\texttt{S\_grad\_abs\_mean}>0.
\]
This reproduces the frozen count of $83{,}088$ active rows.
The Gaussian TRAIN subset contains $58{,}512$ active rows across $1050$
pairs from $350$ sources.
For a controlled paired comparison, all three relationships in
Table~\ref{tab:supp_pseudograd_bootstrap} are therefore evaluated on this same Gaussian TRAIN subset.

\begin{table*}[t]
\centering
\small
\caption{Gaussian TRAIN active-region mechanism analysis with $95\%$ source-bootstrap confidence intervals.}
\label{tab:supp_pseudograd_bootstrap}
\begin{threeparttable}
\begin{tabular}{lcccc}
\toprule
Comparison
& Pearson $r$
& Spearman $\rho_s$
& Sign agreement
& AUROC \\
\midrule
$B_r^{\mathrm{grad}}$ vs. $B_r^{\mathrm{CF}}$
& 0.294 [0.249, 0.339]
& 0.227 [0.212, 0.242]
& 0.594 [0.588, 0.601]
& 0.609 [0.602, 0.615] \\
$B_r^{p}$ vs. $B_r^{\mathrm{CF}}$
& 0.261 [0.213, 0.309]
& 0.144 [0.129, 0.159]
& 0.553 [0.548, 0.559]
& 0.566 [0.559, 0.572] \\
$B_r^{p}$ vs. $B_r^{\mathrm{grad}}$
& 0.767 [0.680, 0.834]
& 0.554 [0.526, 0.581]
& 0.718 [0.705, 0.730]
& 0.752 [0.738, 0.765] \\
\bottomrule
\end{tabular}
\begin{tablenotes}[flushleft]
\footnotesize
\item[] All comparisons use the same $350$ TRAIN sources and $58{,}512$ active
regions with $5000$ source-bootstrap samples. Null baselines are $0$ for
correlations and $0.5$ for sign agreement and AUROC.
\end{tablenotes}
\end{threeparttable}
\end{table*}

The pseudo-gradient has weaker direct agreement with the nonlinear exact
counterfactual than the privileged gradient, but every source-bootstrap
interval remains above the corresponding null or chance baseline.
Moreover, $B_r^p$ remains substantially aligned with
$B_r^{\mathrm{grad}}$, supporting the interpretation that pseudo-labeling
attenuates rather than removes the directional task signal.

\subsection{Gaussian severity diagnostic}
Table~\ref{tab:supp_pseudograd_sigma} further breaks down the deployable
pseudo-gradient agreement with exact counterfactual utility by Gaussian
severity.

\begin{table}[t]
\centering
\small
\caption{$B_r^p$ versus exact $B_r^{\mathrm{CF}}$ by Gaussian severity on active TRAIN regions.}
\label{tab:supp_pseudograd_sigma}
\begin{threeparttable}
\begin{tabular}{rrrrr}
\toprule
$\sigma$ & Pearson & Spearman & Sign agree. & AUROC \\
\midrule
15 & 0.134 & 0.105 & 0.543 & 0.552 \\
25 & 0.214 & 0.111 & 0.544 & 0.554 \\
50 & 0.271 & 0.196 & 0.572 & 0.585 \\
\bottomrule
\end{tabular}
\begin{tablenotes}[flushleft]
\footnotesize
\item[] Diagnostic only; all rows use active Gaussian TRAIN regions.
\end{tablenotes}
\end{threeparttable}
\end{table}

The association becomes stronger at the highest Gaussian severity.
We do not treat this monotonic pattern as a primary claim because the analysis
is restricted to Gaussian TRAIN data.

\section{Threshold Operating-Point Sensitivity}
\label{supp:threshold}

Table~\ref{tab:supp_threshold_points} reports the family of operating points
obtained by applying the same TRAIN-only rule under nearby benefit-retention
constraints.
The final columns are evaluated only after all thresholds have been selected
from TRAIN.

\begin{table*}[t]
\centering
\small
\caption{TRAIN-selected operating points and post-freeze zero-shot sensitivity.}
\label{tab:supp_threshold_points}
\begin{threeparttable}
\begin{tabular*}{\textwidth}{@{\extracolsep{\fill}}rrrrrrrr@{}}
\toprule
TRAIN constraint
& $\tau$
& TRAIN ret.
& TRAIN neg. red.
& TRAIN preserve
& ZS ret.
& ZS neg. red.
& ZS preserve \\
\midrule
99.5\% & 0.0066 & 99.61\% & 28.93\% & 13.05\% & 95.51\% & 41.53\% & 24.17\% \\
99.0\% & 0.0467 & 99.09\% & 33.50\% & 15.90\% & 93.90\% & 45.54\% & 27.00\% \\
98.0\% & 0.1076 & 98.02\% & 39.59\% & 20.19\% & 93.25\% & 54.23\% & 32.72\% \\
97.0\% & 0.1657 & 97.06\% & 46.70\% & 24.57\% & 90.98\% & 60.26\% & 37.39\% \\
95.0\% & 0.2860 & 95.10\% & 57.36\% & 33.14\% & 88.05\% & 70.34\% & 44.78\% \\
\bottomrule
\end{tabular*}
\begin{tablenotes}[flushleft]
\footnotesize
\item[] Ret.: benefit retention; Neg. red.: relative loss-negative reduction;
ZS: pooled zero-shot motion, rain, and defocus.
\end{tablenotes}
\end{threeparttable}
\end{table*}

The post-freeze zero-shot sensitivity is descriptive only.
It demonstrates that the frozen $98\%$-retention point lies on a smooth
risk--utility tradeoff rather than being an isolated test-selected threshold.

\section{Appearance-Baseline Details}
\label{supp:appearance}

The Image-only baseline uses frozen ResNet-50 features from the degraded image.
The Visual-pair baseline uses
\[
[f_d,f_c,f_c-f_d,|f_c-f_d|].
\]
Both follow the same Gaussian-only fitting and freezing discipline as
TaskGuard.

On unseen motion/rain/defocus, Image-only obtains
$\rho_s=0.091$, benefit retention $86.1\%$, and loss-negative reduction
$6.9\%$.
Visual-pair increases utility-ranking correlation to
$\rho_s=0.211$ and loss-negative reduction to $30.3\%$, while its benefit
retention is $79.6\%$.
TaskGuard reaches
$\rho_s=0.622$, benefit retention $93.3\%$, and loss-negative reduction
$54.2\%$.
The source-bootstrap improvements of TaskGuard over Visual-pair are
$+19.5$ percentage points in benefit retention
(95\% CI $[14.2,25.1]$),
$+33.6$ points in loss-negative reduction
($[25.6,41.8]$),
and $+15.2$ points in practical-negative reduction
($[7.7,22.8]$).

\section{Natural-Rain DAWN Details}
\label{supp:dawn}

DAWN contains $198$ natural-rain images in the evaluation used here.
Original rainy images obtain COCO-style AP $0.357807$ and unconditional
deraining obtains $0.381787$.

The Image-only policy reaches AP $0.380475$, restores $189$ of $198$ images,
retains $94.5\%$ of the Always-Derain AP gain, retains $99.7\%$ of
loss-defined restoration benefit, and reduces loss-negative interventions by
$6.4\%$.

Visual-pair reaches AP $0.377305$, restores $41$ images, retains $81.3\%$ of
the AP gain, retains $101.2\%$ of loss-defined benefit, and reduces
loss-negative interventions by $80.9\%$.

TaskGuard reaches AP $0.376466$, restores only $8$ images, retains $77.8\%$ of
the AP gain, retains $100.4\%$ of loss-defined benefit, and reduces
loss-negative interventions by $97.9\%$.
Benefit retention can exceed $100\%$ because preserving selected
negative-utility candidates can yield greater mean policy utility than
unconditional deraining.
Only six Always-Derain samples are practical-negative, so the corresponding
practical-risk comparisons are descriptive.

Table~\ref{tab:supp_dawn_ap_ci} reports paired source-bootstrap AP
differences between TaskGuard and the comparison policies.

\begin{table}[t]
\centering
\small
\caption{Paired source-bootstrap DAWN AP differences.}
\label{tab:supp_dawn_ap_ci}
\begin{threeparttable}
\begin{tabular}{lrr}
\toprule
Comparison & AP diff. & 95\% CI \\
\midrule
TaskGuard $-$ Visual-pair
& -0.00084 & [-0.00277, 0.00098] \\
TaskGuard $-$ Image-only
& -0.00401 & [-0.01248, 0.00266] \\
TaskGuard $-$ Always Derain
& -0.00532 & [-0.01332, 0.00208] \\
\bottomrule
\end{tabular}
\begin{tablenotes}[flushleft]
\footnotesize
\item[] All intervals are paired $95\%$ source-bootstrap CIs and include zero;
they do not establish equivalence or non-inferiority.
\end{tablenotes}
\end{threeparttable}
\end{table}

TaskGuard improves loss-negative reduction over Visual-pair by
$17.0$ percentage points with paired source-bootstrap
95\% CI $[8.4,25.8]$.

\begin{figure}[t]
    \centering
    \includegraphics[width=\linewidth]
    {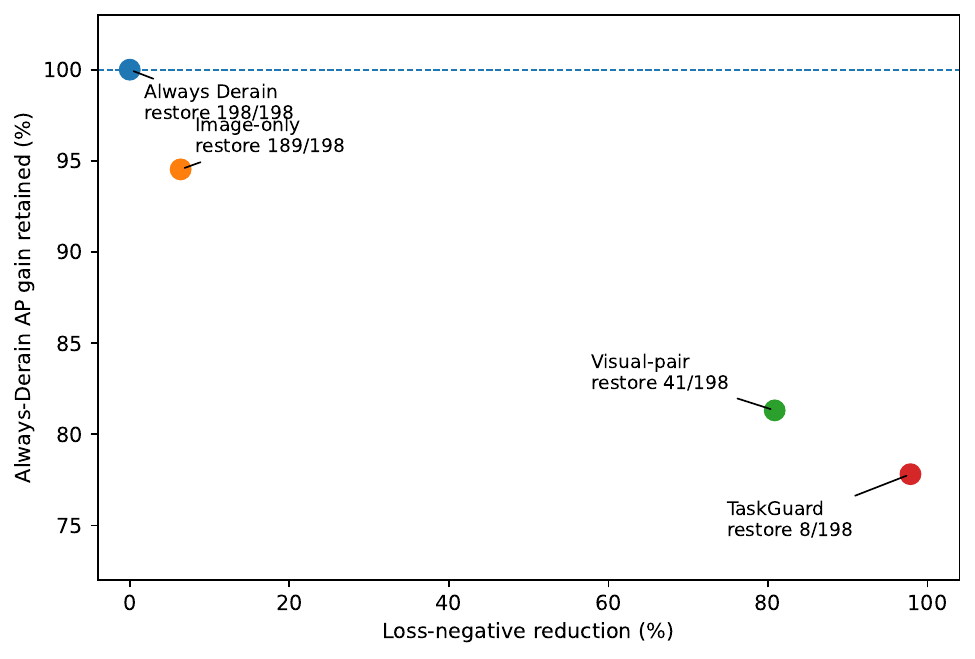}
    \caption{Natural-rain DAWN risk--accuracy tradeoff.}
    \label{fig:supp_dawn_tradeoff}
\end{figure}

Figure~\ref{fig:supp_dawn_tradeoff} visualizes the corresponding
risk--accuracy tradeoff on natural-rain DAWN.

\section{Cross-Detector Safety Transfer}
\label{supp:cross_detector}

The TaskGuard score and decision are constructed using YOLO11s.
To test whether the decision transfers to another detector-specific utility
function, we reevaluate the same frozen decisions on RT-DETR-L without
retraining, changing features, or retuning $\tau$.

On the $600$ fresh Gaussian pairs, the frozen TaskGuard score has
$\rho_s=0.3048$ with RT-DETR-L restoration utility.
The policy retains $93.92\%$ of RT-DETR-L loss-defined benefit and reduces the
RT-DETR-L loss-negative rate from $28.50\%$ to $19.83\%$, a relative reduction
of $30.41\%$.

For comparison, under the YOLO11s utility used to construct TaskGuard, the
fresh Gaussian result is $\rho_s=0.7260$, benefit retention $96.98\%$, and a
loss-negative-rate reduction from $19.50\%$ to $11.00\%$.
The weaker RT-DETR result motivates the main-paper wording
\emph{partial cross-detector safety transfer} rather than detector invariance.

\section{Practical Detection-Utility Analysis}
\label{supp:practical}

For each degraded/restored pair we compute class-aware matches at IoU $0.50$
and detector confidence $0.25$.
Let
\[
\Delta F1=F1_c-F1_d,\qquad
\Delta FN=FN_c-FN_d.
\]
A restoration is practical-negative when
\[
\Delta F1<0
\quad\text{or}\quad
\Delta FN>0.
\]

Across all controlled final pairs, the loss-negative rate is $21.12\%$ and the
practical-negative rate is $18.08\%$.
The association between whole-image loss utility and per-image F1 change is
\[
\rho_s(B_{\mathrm{full}},\Delta F1)=0.601,
\]
and
$-B_{\mathrm{full}}$
identifies practical-negative cases with AUROC $0.700$.

A practical deterioration occurs in $34.32\%$ of loss-negative restorations but
only $13.73\%$ of non-loss-negative restorations.
On the pooled unseen families, Always Restore is practical-negative on
$18.33\%$ of pairs and TaskGuard on $11.56\%$, a relative reduction of
$36.97\%$.

These results support the use of detector loss as a differentiable utility
proxy while also showing why it should not be treated as identical to
task-level detection quality.

\section{Runtime Details}
\label{supp:runtime}

\runinhead{Runtime protocol.}
Runtime was measured on a single NVIDIA RTX A4000 GPU using YOLO11s at
input size $640$.
The benchmark uses $50$ Gaussian validation sources at $\sigma=25$, with
$5$ warm-up samples before timing.
The benchmark was invoked on \texttt{cuda:0}, and CUDA synchronization is
performed immediately before and after each timed GPU operation.
The timing experiment uses development data only and does not access the
fresh final benchmark.

Table~\ref{tab:supp_runtime} summarizes the measured TaskGuard decision-stage
runtime and the contextual end-to-end pipeline timings.

\begin{table}[t]
\centering
\caption{Runtime of the frozen TaskGuard decision stage.}
\label{tab:supp_runtime}
\begin{threeparttable}
\small
\setlength{\tabcolsep}{4.5pt}
\begin{tabular}{lrr}
\toprule
Configuration & Time / image & Increment \\
\midrule
Detector only & 26.7 ms & -- \\
TaskGuard post-restoration & 162.9 ms & +136.1 ms \\
\midrule
Restore + detector & 667 ms & contextual \\
Restore + TaskGuard & 803 ms & contextual \\
\bottomrule
\end{tabular}
\begin{tablenotes}[flushleft]
\footnotesize
\item[] Core timings are measured after a restored candidate is available.
Contextual rows add the saved Restormer directory timing, including startup.
TaskGuard adds $136.1$ ms/image over detector-only inference, or approximately
$20.4\%$ relative to Restore+Detector.
\end{tablenotes}
\end{threeparttable}
\end{table}

\runinhead{Timing interpretation.}
The measured detector-only time is $26.731$ ms/image.
The complete TaskGuard post-restoration stage requires $162.865$ ms/image,
corresponding to an incremental TaskGuard cost of $136.134$ ms/image.

The separately measured Restormer directory run requires $640.310$ ms/image
and includes Python process and model-initialization overhead.
Using this contextual restoration timing gives approximately
$667.041$ ms/image for Restore+Detector and $803.175$ ms/image for
Restore+TaskGuard.
Thus, the TaskGuard increment is approximately $20.4\%$ relative to the
measured Restore+Detector pipeline.
Because the Restormer measurement includes process startup and model
initialization, these combined numbers are provided as deployment context
rather than optimized persistent-model latency.

\section{Reproducibility Checklist}
\label{supp:reproducibility}

The final evaluation follows the following lock discipline:

\begin{enumerate}
    
    \item All controlled source images are drawn from COCO 2017
    \texttt{val2017}.

    \item Gaussian TRAIN contains $350$ sources and $1050$ pairs.

    \item The frozen TaskGuard representation has exactly $43$ features.

    \item The frozen candidate is
    \texttt{StandardScaler + Ridge(alpha=10)}.

    \item The inference score is the average of $50$ fold-specific regressors
    from ten repeated five-fold source-grouped stratified splits.

    \item The frozen threshold is
    $\tau=0.107633802381$.

    \item Threshold selection uses Gaussian TRAIN OOF predictions only.

    \item The final controlled benchmark contains $200$ untouched,
    source-disjoint sources and $2400$ pairs.

    \item Motion blur, rain, and defocus are unseen during TaskGuard fitting.

    \item No final controlled, DAWN, or RT-DETR result is used for feature,
    model, or threshold retuning.

    \item Post-freeze feature ablations, pseudo-gradient diagnostics, and
    threshold-sensitivity analyses do not change the locked TaskGuard policy.
\end{enumerate}


%% file: main.bib
@String(ECCV= {Eur. Conf. Comput. Vis.})

@String(AAAI = {AAAI})

@String(ECCV  = {ECCV})

@inproceedings{zhao2024rtdetr,
  title     = {{DETRs Beat YOLOs on Real-time Object Detection}},
  author    = {Zhao, Yian and Lv, Wenyu and Xu, Shangliang and Wei, Jinman
               and Wang, Guanzhong and Dang, Qingqing and Liu, Yi and Chen, Jie},
  booktitle = {Proceedings of the IEEE/CVF Conference on Computer Vision and Pattern Recognition},
  pages     = {16965--16974},
  year      = {2024},
  doi       = {10.1109/CVPR52733.2024.01605}
}

@inproceedings{liu2022iayolo,
  title     = {Image-Adaptive YOLO for Object Detection in Adverse Weather Conditions},
  author    = {Liu, Wenyu and Ren, Gaofeng and Yu, Runsheng and Guo, Shi and Zhu, Jianke and Zhang, Lei},
  booktitle = {Proceedings of the AAAI Conference on Artificial Intelligence},
  volume    = {36},
  number    = {2},
  pages     = {1792--1800},
  year      = {2022},
  doi       = {10.1609/aaai.v36i2.20072}
}

@article{kenk2020dawn,
  title   = {{DAWN}: Vehicle Detection in Adverse Weather Nature Dataset},
  author  = {Kenk, Mourad A. and Hassaballah, Mahmoud},
  journal = {arXiv preprint arXiv:2008.05402},
  year    = {2020}
}

@inproceedings{kalwar2023gdip,
  title     = {{GDIP}: Gated Differentiable Image Processing for Object Detection in Adverse Conditions},
  author    = {Kalwar, Sanket and Patel, Dhruv and Aanegola, Aakash and Konda, Krishna Reddy and Garg, Sourav and Krishna, K. Madhava},
  booktitle = {2023 IEEE International Conference on Robotics and Automation (ICRA)},
  pages     = {7083--7089},
  year      = {2023},
  doi       = {10.1109/ICRA48891.2023.10160356}
}

@article{li2023detectionfriendly,
  title   = {Detection-Friendly Dehazing: Object Detection in Real-World Hazy Scenes},
  author  = {Li, Chengyang and Zhou, Heng and Liu, Yang and Yang, Caidong and Xie, Yongqiang and Li, Zhongbo and Zhu, Liping},
  journal = {IEEE Transactions on Pattern Analysis and Machine Intelligence},
  volume  = {45},
  number  = {7},
  pages   = {8284--8295},
  year    = {2023},
  doi     = {10.1109/TPAMI.2023.3234976}
}

@inproceedings{kim2024sr4ir,
  title     = {Beyond Image Super-Resolution for Image Recognition with Task-Driven Perceptual Loss},
  author    = {Kim, Jaeha and Oh, Junghun and Lee, Kyoung Mu},
  booktitle = {Proceedings of the IEEE/CVF Conference on Computer Vision and Pattern Recognition},
  pages     = {2651--2661},
  year      = {2024},
  doi       = {10.1109/CVPR52733.2024.00256}
}

@inproceedings{chen2025unirestore,
  title     = {UniRestore: Unified Perceptual and Task-Oriented Image Restoration Model Using Diffusion Prior},
  author    = {Chen, I-Hsiang and Chen, Wei-Ting and Liu, Yu-Wei and Chiang, Yuan-Chun and Kuo, Sy-Yen and Yang, Ming-Hsuan},
  booktitle = {Proceedings of the IEEE/CVF Conference on Computer Vision and Pattern Recognition},
  pages     = {17969--17979},
  year      = {2025},
  doi       = {10.1109/CVPR52734.2025.01674}
}

@inproceedings{kim2025edtr,
  title     = {Exploiting Diffusion Prior for Task-Driven Image Restoration},
  author    = {Kim, Jaeha and Oh, Junghun and Lee, Kyoung Mu},
  booktitle = {Proceedings of the IEEE/CVF International Conference on Computer Vision},
  pages     = {10151--10161},
  year      = {2025},
  doi       = {10.1109/ICCV51701.2025.00946}
}

@inproceedings{li2025machinepreference,
  title     = {Image Quality Assessment: From Human to Machine Preference},
  author    = {Li, Chunyi and Tian, Yuan and Ling, Xiaoyue and Zhang, Zicheng and Duan, Haodong and Wu, Haoning and Jia, Ziheng and Liu, Xiaohong and Min, Xiongkuo and Lu, Guo and Lin, Weisi and Zhai, Guangtao},
  booktitle = {Proceedings of the IEEE/CVF Conference on Computer Vision and Pattern Recognition},
  pages     = {7570--7581},
  year      = {2025},
  doi       = {10.1109/CVPR52734.2025.00709}
}

@article{fang2026lqtaranker,
  title   = {{LQTARanker}: A Ranking-Based Task-Oriented Assessment Framework for Low-Quality Images},
  author  = {Fang, Shengyu and Guo, Jichang},
  journal = {Neurocomputing},
  volume  = {704},
  pages   = {134815},
  year    = {2026},
  doi     = {10.1016/j.neucom.2026.134815}
}

@inproceedings{lee2026tasktok,
  title     = {TaskTok: Delving into Task Tokens for Task-driven Image Restoration},
  author    = {Lee, Hongjae and Kang, Sojung and Yu, Jaeseong and Jung, Seung-Won},
  booktitle = {European Conference on Computer Vision},
  year      = {2026}
}

@inproceedings{lin2014coco,
  title     = {Microsoft {COCO}: Common Objects in Context},
  author    = {Lin, Tsung-Yi and Maire, Michael and Belongie, Serge
               and Hays, James and Perona, Pietro and Ramanan, Deva
               and Doll{\'a}r, Piotr and Zitnick, C. Lawrence},
  booktitle = {Computer Vision -- ECCV 2014},
  pages     = {740--755},
  year      = {2014},
  doi       = {10.1007/978-3-319-10602-1_48}
}

@inproceedings{zamir2022restormer,
  title     = {Restormer: Efficient Transformer for High-Resolution Image Restoration},
  author    = {Zamir, Syed Waqas and Arora, Aditya and Khan, Salman
               and Hayat, Munawar and Khan, Fahad Shahbaz
               and Yang, Ming-Hsuan},
  booktitle = {Proceedings of the IEEE/CVF Conference on Computer Vision and Pattern Recognition},
  pages     = {5728--5739},
  year      = {2022}
}

@misc{yolo11_ultralytics,
  author  = {Glenn Jocher and Jing Qiu},
  title   = {Ultralytics YOLO11},
  version = {11.0.0},
  year    = {2024}
}

@inproceedings{he2016resnet,
  title     = {Deep Residual Learning for Image Recognition},
  author    = {He, Kaiming and Zhang, Xiangyu and Ren, Shaoqing and Sun, Jian},
  booktitle = {Proceedings of the IEEE Conference on Computer Vision and Pattern Recognition},
  pages     = {770--778},
  year      = {2016},
  doi       = {10.1109/CVPR.2016.90}
}
